\documentclass{article}

\usepackage{times}
\usepackage{graphicx}
\usepackage{subfigure}

\usepackage{natbib}
\usepackage{algorithm}
\usepackage{algorithmic}

\usepackage{hyperref}

\usepackage{amsmath}
\usepackage{amssymb}
\usepackage{ifthen}
\usepackage[utf8]{inputenc}
\usepackage{url}

\usepackage[accepted]{icml2017}

\newcommand{\twonorm}[1]{\left\lVert#1\right\rVert_2^2}

\newcommand\theTitle{Understanding Black-box Predictions via Influence Functions}

\icmltitlerunning{\theTitle}

\newcommand\sI{\ensuremath{\mathcal{I}}}

\newcommand\sX{\ensuremath{\mathcal{X}}}
\newcommand\sY{\ensuremath{\mathcal{Y}}}

\DeclareMathOperator*{\sign}{sign}
\newcommand\pb[1]{\ensuremath{\left[ #1 \right]}} 
\newcommand\pc[1]{\ensuremath{\left\{ #1 \right\}}} 
\newcommand\inv[1]{\ensuremath{\frac{1}{#1}}}

\newcommand\eqdef{\ensuremath{\stackrel{\rm def}{=}}} 
\newcommand\refeqn[1]{(\ref{eqn:#1})}

\ifthenelse{\isundefined{\definition}}{\newtheorem{definition}{Definition}}{}
\ifthenelse{\isundefined{\assumption}}{\newtheorem{assumption}{Assumption}}{}
\ifthenelse{\isundefined{\hypothesis}}{\newtheorem{hypothesis}{Hypothesis}}{}
\ifthenelse{\isundefined{\proposition}}{\newtheorem{proposition}{Proposition}}{}
\ifthenelse{\isundefined{\theorem}}{\newtheorem{theorem}{Theorem}}{}
\ifthenelse{\isundefined{\lemma}}{\newtheorem{lemma}{Lemma}}{}
\ifthenelse{\isundefined{\corollary}}{\newtheorem{corollary}{Corollary}}{}
\ifthenelse{\isundefined{\alg}}{\newtheorem{alg}{Algorithm}}{}
\ifthenelse{\isundefined{\example}}{\newtheorem{example}{Example}}{}
\newcommand\cv{\ensuremath{\to}} 

\newcommand\inflparams[1]{\sI_\text{up,params}#1}
\newcommand\inflloss[1]{\sI_\text{up,loss}#1}
\newcommand\inflinput[1]{\sI_\text{pert,loss}#1}

\begin{document}

\twocolumn[
\icmltitle{\theTitle}
\icmlsetsymbol{equal}{*}
\vspace{-2mm}

\begin{icmlauthorlist}
\icmlauthor{Pang Wei Koh}{stanford}
\icmlauthor{Percy Liang}{stanford}
\end{icmlauthorlist}

\icmlcorrespondingauthor{Pang Wei Koh}{pangwei@cs.stanford.edu}
\icmlcorrespondingauthor{Percy Liang}{pliang@cs.stanford.edu}

\icmlaffiliation{stanford}{Stanford University, Stanford, CA}

\icmlkeywords{influence}

\vskip 0.2in
]
\printAffiliationsAndNotice{}

\begin{abstract}
How can we explain the predictions of a black-box model?
In this paper, we use influence functions --- a classic technique from robust statistics --- to trace a model's prediction through the learning algorithm and back to its training data, thereby identifying training points most responsible for a given prediction.
To scale up influence functions to modern machine learning settings,
we develop a simple, efficient implementation that requires only oracle access to gradients and Hessian-vector products.
We show that even on non-convex and non-differentiable models
where the theory breaks down,
approximations to influence functions can still provide valuable information.
On linear models and convolutional neural networks,
we demonstrate that influence functions are useful for multiple purposes:
understanding model behavior, debugging models, detecting dataset errors,
and even creating visually-indistinguishable training-set attacks.

\end{abstract}
\vspace{-6mm}

\section{Introduction}

A key question often asked of machine learning systems is ``Why did the system make this prediction?"
We want models that are not just high-performing but also explainable.
By understanding why a model does what it does, we can hope to improve the model \citep{amershi2015modeltracker}, discover new science \citep{shrikumar2017learning}, and provide end-users with explanations of actions that impact them \citep{goodman2016eu}.

However, the best-performing models in many domains --- e.g., deep neural networks for image and speech recognition \citep{krizhevsky2012imagenet} --- are complicated, black-box models whose predictions seem hard to explain. Work on interpreting these black-box models has focused on understanding how a fixed model leads to particular predictions, e.g., by locally fitting a simpler model around the test point \citep{ribeiro2016lime} or by perturbing the test point to see how the prediction changes \citep{simonyan2013deep, li2016understanding, datta2016algorithmic, adler2016auditing}. These works explain the predictions in terms of the model, but how can we explain where the model came from?

In this paper, we tackle this question by tracing a model's predictions through its learning algorithm and back to the training data, where the model parameters ultimately derive from.
To formalize the impact of a training point on a prediction, we ask the counterfactual: what would happen if we did not have this training point, or if the values of this training point were changed slightly?

Answering this question by perturbing the data and retraining the model can be prohibitively expensive.
To overcome this problem, we use influence functions, a classic technique from robust statistics \citep{hampel1974influence} that tells us how the model parameters change as we upweight a training point by an infinitesimal amount.
This allows us to
``differentiate through the training'' to
estimate in closed-form the effect of a variety of training perturbations.

Despite their rich history in statistics, influence functions have not seen widespread use in machine learning; to the best of our knowledge, the work closest to ours is \citet{wojnowicz2016sketching}, which introduced a method for approximating
a quantity related to influence
in generalized linear models. One obstacle to adoption is that influence functions require expensive second derivative calculations and assume model differentiability and convexity, which limits their applicability in modern contexts where models are often non-differentiable, non-convex, and high-dimensional.
We address these challenges by showing that we can efficiently approximate influence functions using second-order optimization techniques \citep{pearlmutter1994fast, martens2010deep, agarwal2016second}, and that they remain accurate even as the underlying assumptions of differentiability and convexity degrade.

Influence functions capture the core idea of studying models through the lens of their training data. We show that they are a versatile tool that can be applied to a wide variety of seemingly disparate tasks: understanding model behavior, debugging models, detecting dataset errors, and creating visually-indistinguishable adversarial \emph{training} examples that can flip neural network test predictions, the training set analogue of \citet{goodfellow2015explaining}.

\section{Approach}
\label{sec:approach}

Consider a prediction problem from some input space $\sX$ (e.g., images) to an
output space $\sY$ (e.g., labels).
We are given training points
$z_1, \ldots, z_n$, where $z_i = (x_i, y_i) \in \mathcal{X} \times \mathcal{Y}$.
For a point $z$ and parameters $\theta \in \Theta$,
let $L(z, \theta)$ be the loss,
and let $\frac{1}{n} \sum_{i=1}^n L(z_i, \theta)$ be the empirical risk.
The empirical risk minimizer is given by
$\hat\theta \eqdef \arg\min_{\theta\in\Theta} \frac{1}{n}
\sum_{i=1}^n L(z_i,\theta)$.\footnote{We fold in any regularization terms into $L$.}
Assume that the empirical risk
is twice-differentiable and strictly convex in $\theta$;
in Section \ref{sec:approx} we explore relaxing these assumptions.

\subsection{Upweighting a training point}

Our goal is to understand the effect of training points on a model's predictions. We formalize this goal by asking the counterfactual: how would the model's predictions change if we did not have this training point?

Let us begin by studying the change in model parameters due to removing a point $z$ from the training set.
Formally, this change is $\hat\theta_{-z} - \hat\theta$,
where $\hat\theta_{-z} \eqdef \arg\min_{\theta\in\Theta} \sum_{z_i \neq z} L(z_i,\theta)$.
However, retraining the model for each removed $z$ is prohibitively slow.

Fortunately, influence functions give us an efficient approximation. The idea
is to compute the parameter change if $z$ were upweighted by some small
$\epsilon$, giving us new parameters $\hat\theta_{\epsilon, z} \eqdef
\arg\min_{\theta\in\Theta} \frac{1}{n} \sum_{i=1}^n L(z_i,\theta)
+ \epsilon L(z, \theta)$. A classic result \citep{cook1982residuals} tells us
that the influence of upweighting $z$ on the parameters $\hat\theta$ is given
by
\begin{align}
\label{eqn:params}
\inflparams(z) & \eqdef \frac{d\hat\theta_{\epsilon, z}}{d\epsilon}\Bigr|_{\epsilon = 0} = -H_{\hat\theta}^{-1} \ \nabla_\theta L(z, \hat\theta),
\end{align}
where $H_{\hat\theta} \eqdef \frac{1}{n} \sum_{i=1}^n \nabla^2_\theta L(z_i, \hat\theta)$ is the Hessian and is positive definite (PD) by assumption.
In essence, we are forming a quadratic approximation to the empirical risk around $\hat\theta$ and take a single Newton step; see appendix A for a derivation.
Since removing a point $z$ is the same as upweighting it by $\epsilon = -\frac{1}{n}$,
we can linearly approximate the parameter change due to removing $z$ without retraining the model by computing
$\hat\theta_{-z} - \hat\theta \approx -\frac{1}{n} \inflparams(z)$.

Next, we apply the chain rule to measure how upweighting $z$ changes functions of $\hat\theta$. In particular, the influence of upweighting $z$ on the loss at a test point $z_\text{test}$
again has a closed-form expression:
\begin{align}
\label{eqn:influence}
\inflloss(z, z_\text{test}) & \eqdef \frac{dL(z_\text{test}, \hat\theta_{\epsilon, z})}{d\epsilon}\Bigr|_{\substack{\epsilon = 0}} \\
& = \nabla_\theta L(z_\text{test}, \hat\theta) ^\top \frac{d\hat\theta_{\epsilon, z}}{d\epsilon}\Bigr|_{\substack{\epsilon = 0}} \nonumber \\
& = -\nabla_\theta L(z_\text{test}, \hat\theta) ^\top H_{\hat\theta}^{-1} \nabla_\theta L(z, \hat\theta). \nonumber
\end{align}

\subsection{Perturbing a training input}
\label{sec:approach-input}

Let us develop a finer-grained notion of influence by studying a different counterfactual: how would the model's predictions change if a training input were modified?

For a training point $z=(x,y)$,
define $z_\delta \eqdef (x + \delta, y)$.
Consider the perturbation $z \mapsto z_\delta$, and let $\hat\theta_{z_\delta, -z}$ be
the empirical risk minimizer on the training points with $z_\delta$ in place of $z$.
To approximate its effects,
define the parameters resulting from moving $\epsilon$ mass from $z$ onto $z_\delta$:
$\hat\theta_{\epsilon,z_\delta,-z} \eqdef \arg\min_{\theta \in \Theta} \inv{n} \sum_{i=1}^n L(z_i, \theta) + \epsilon L(z_\delta, \theta) - \epsilon L(z, \theta)$.
An analogous calculation to \refeqn{params} yields:
\begin{align}
\label{eqn:inflinput-discrete}
\frac{d\hat\theta_{\epsilon, z_\delta, -z}}{d\epsilon}\Bigr|_{\substack{\epsilon = 0}} &= \inflparams(z_\delta) -\inflparams(z) \nonumber\\
&= -H_{\hat\theta}^{-1} \big(\nabla_\theta L(z_\delta, \hat\theta) - \nabla_\theta L(z, \hat\theta) \big).
\end{align}
As before, we can make the linear approximation
$\hat\theta_{z_\delta, -z} - \hat\theta \approx
\frac{1}{n}(\inflparams(z_\delta) -\inflparams(z))$,
giving us a closed-form estimate of the effect of $z \mapsto z_\delta$ on the model.
Analogous equations also apply for changes in $y$.
While influence functions might appear to only work for infinitesimal (therefore continuous) perturbations,
it is important to note that this approximation holds for arbitrary $\delta$: the $\epsilon$-upweighting scheme allows us to smoothly interpolate between $z$ and $z_\delta$. This is particularly useful for working with discrete data (e.g., in NLP) or with discrete label changes.

If $x$ is continuous and $\delta$ is small, we can further approximate \refeqn{inflinput-discrete}. Assume that the input domain $\mathcal{X} \subseteq \mathbb{R}^d$, the parameters $\Theta \subseteq \mathbb{R}^p$, and $L$ is differentiable in $\theta$ and $x$. As $\|\delta\| \rightarrow 0$, $\nabla_\theta L(z_\delta, \hat\theta) - \nabla_\theta L(z, \hat\theta) \approx [\nabla_x \nabla_\theta L(z, \hat\theta)]\delta$, where $\nabla_x \nabla_\theta L(z, \hat\theta) \in \mathbb{R}^{p \times d}$. Substituting into \refeqn{inflinput-discrete},
\begin{align}
\frac{d\hat\theta_{\epsilon, z_\delta, -z}}{d\epsilon}\Bigr|_{\substack{\epsilon = 0}} \approx -H_{\hat\theta}^{-1}  [\nabla_x \nabla_\theta L(z, \hat\theta)]\delta.
\end{align}
We thus have $\hat\theta_{z_\delta, -z} - \hat\theta \approx -\frac{1}{n} H_{\hat\theta}^{-1}  [\nabla_x \nabla_\theta L(z, \hat\theta)]\delta$. Differentiating w.r.t. $\delta$ and applying the chain rule gives us
\begin{align}
\label{eqn:delta_x_loss}
\inflinput(z, z_\text{test}) & \eqdef \nabla_\delta L(z_\text{test}, \hat\theta_{z_\delta, -z}) \Bigr|_{\substack{\delta = 0}}\\
& = -\nabla_\theta L(z_\text{test}, \hat\theta)^\top H_{\hat\theta}^{-1} \nabla_x \nabla_\theta L(z, \hat\theta).\nonumber
\end{align}
$[\inflinput(z, z_\text{test})] \delta$ tells us the approximate effect that $z \mapsto z + \delta$ has on the loss at $z_\text{test}$. By setting $\delta$ in the direction of $\inflinput(z, z_\text{test})^\top$, we can construct local perturbations of $z$ that maximally increase the loss at $z_\text{test}$. In Section \ref{sec:dataset-poisoning}, we will use this to construct training-set attacks. Finally, we note that $\inflinput(z, z_\text{test})$ can help us identify the features of $z$ that are most responsible for the prediction on $z_\text{test}$.

\begin{figure*}[!t]
\centerline{
\includegraphics[width=\textwidth]{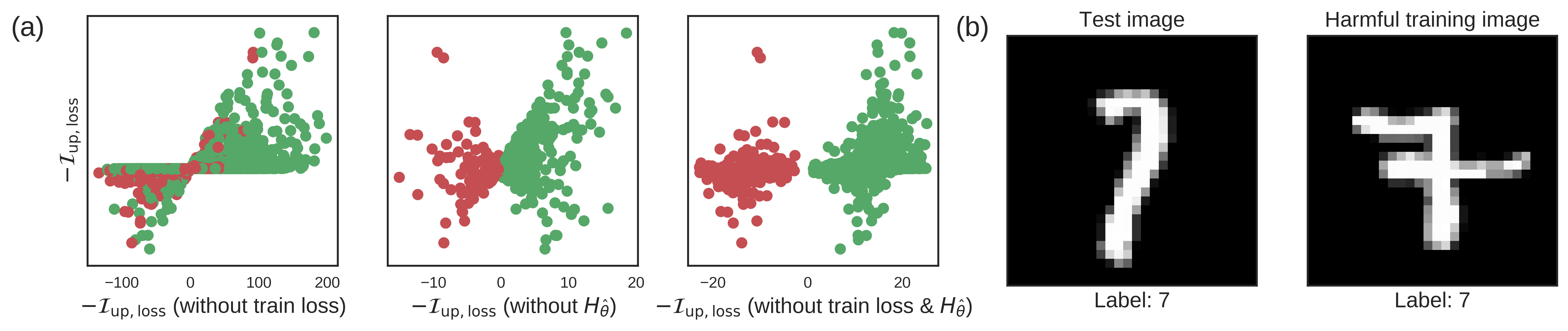}}
\vspace{-2mm}
\caption{{\bf Components of influence.} {\bf (a)} What is the effect of the training loss and $H_{\hat\theta}^{-1}$ terms in $\inflloss$? Here, we plot $\inflloss$ against variants that are missing these terms and show that they are necessary for picking up the truly influential training points. For these calculations, we use logistic regression to distinguish 1's from 7's in MNIST \citep{lecun1998gradient}, picking an arbitrary test point $z_\text{test}$; similar trends hold across other test points. Green dots are train images of the same label as the test image (7) while red dots are 1's.
  {\bf Left:} Without the train loss term, we overestimate the influence of many training points: the points near the horizontal axis should have $\inflloss$ close to 0, but instead have high influence when we remove the train loss term). {\bf Mid:} Without $H_{\hat\theta}^{-1}$, all green training points are helpful (removing each point increases test loss) and all red points are harmful (removing each point decreases test loss). This is because $\forall x, x \succeq 0$ (all pixel values are positive), so $x \cdot x_\text{test} \geq 0$, but it is incorrect: many harmful training points actually share the same label as $z_\text{test}$. See panel (b). {\bf Right:} Without training loss or $H_{\hat\theta}^{-1}$, what is left is the scaled Euclidean inner product $y_\text{test} y \cdot \sigma(-y_\text{test} \theta^\top \cdot x_\text{test}) x_\text{test}^\top x$, which fails to accurately capture influence; the scatter plot deviates quite far from the diagonal. {\bf (b)} The test image and a harmful training image with the same label. To the model, they look very different, so the presence of the training image makes the model think that the test image is less likely to be a 7. The Euclidean inner product does not pick up on these less intuitive, but important, harmful influences.
}
\vspace{-2mm}
\label{fig:knn-comparison}
\end{figure*}

\subsection{Relation to Euclidean distance}
\label{sec:knn}
To find the training points most relevant to a test point, it is common to look at its nearest neighbors in Euclidean space (e.g., \citet{ribeiro2016lime}); if all points have the same norm, this is equivalent to choosing $x$ with the largest $x \cdot x_\text{test}$. For intuition, we compare this to $\inflloss(z, z_\text{test})$ on a logistic regression model and show that influence is much more accurate at accounting for the effect of training.

Let $p(y \mid x) = \sigma(y \theta^\top x)$, with $y \in \{-1, 1\}$ and $\sigma(t) = \frac{1}{1 + \exp(-t)}$. We seek to maximize the probability of the training set. For a training point $z=(x,y)$,
$L(z, \theta) = \log (1 + \exp(-y \theta^\top x))$, $\nabla_\theta L(z, \theta) = -\sigma(-y \theta^\top x) yx$, and $H_\theta = \frac{1}{n} \sum_{i=1}^n \sigma(\theta^\top x_i) \sigma(-\theta^\top x_i) x_i x_i^\top$.
From \refeqn{influence}, $\inflloss(z, z_\text{test})$ is:
\begin{align*}
-y_\text{test} y \cdot \sigma(-y_\text{test} \theta^\top x_\text{test}) \cdot \sigma(-y \theta^\top x) \cdot x_\text{test}^\top H_{\hat\theta}^{-1} x.
\end{align*}
We highlight two key differences from $x \cdot x_\text{test}$.
First, $\sigma(-y \theta^\top x)$ gives points with high training loss more influence, revealing that outliers can dominate the model parameters. Second, the weighted covariance matrix $H_{\hat\theta}^{-1}$ measures the ``resistance'' of the other training points to the removal of $z$; if $\nabla_\theta L(z, \hat\theta)$ points in a direction of little variation, its influence will be higher since moving in that direction will not significantly increase the loss on other training points. As we show in Fig~\ref{fig:knn-comparison}, these differences mean that influence functions capture the effect of model training much more accurately than nearest neighbors.

\section{Efficiently calculating influence}
\label{sec:efficient}
There are two challenges to efficiently computing $\inflloss(z, z_\text{test})=-\nabla_\theta L(z_\text{test}, \hat\theta) ^\top H_{\hat\theta}^{-1} \nabla_\theta L(z, \hat\theta)$. First, it requires forming and inverting $H_{\hat\theta} = \frac{1}{n} \sum_{i=1}^n \nabla^2_\theta L(z_i, \hat\theta)$, the Hessian of the empirical risk.
With $n$ training points and $\theta \in \mathbb{R}^p$, this requires $O(np^2 + p^3)$ operations, which is too expensive for models like deep neural networks with millions of parameters. Second, we need to calculate $\inflloss(z_i, z_\text{test})$ across all training points $z_i$.

The first problem is well-studied in second-order optimization. The idea is to avoid explicitly computing $H_{\hat\theta}^{-1}$; instead, we use implicit Hessian-vector products (HVPs) to efficiently approximate $s_\text{test} \eqdef H_{\hat\theta}^{-1} \ \nabla_\theta L(z_\text{test}, \hat\theta)$ and then compute $\inflloss(z, z_\text{test}) = -s_\text{test} \cdot \nabla_\theta L(z, \hat\theta)$. This also solves the second problem: for each test point of interest, we can precompute $s_\text{test}$ and then efficiently compute $-s_\text{test} \cdot \nabla_\theta L(z_i, \hat\theta)$ for each training point $z_i$.

We discuss two techniques for approximating $s_\text{test}$, both relying on the fact that the HVP of a single term in $H_{\hat\theta}$, $[\nabla^2_\theta L(z_i, \hat\theta)]v$, can be computed for arbitrary $v$
in the same time that $\nabla_\theta L(z_i, \hat\theta)$ would take, which is typically $O(p)$
\citep{pearlmutter1994fast}.

\textbf{Conjugate gradients (CG).} The first technique is a standard transformation of matrix inversion into an optimization problem. Since $H_{\hat\theta} \succ 0$ by assumption, $H_{\hat\theta}^{-1}v \equiv \arg\min_t \{t^\top H_{\hat\theta} t - v^\top t\}$. We can solve this with CG approaches that only require the evaluation of $H_{\hat\theta}t$, which takes $O(np)$ time, without explicitly forming $H_{\hat\theta}$. While an exact solution takes $p$ CG iterations, in practice we can get a good approximation with fewer iterations; see \citet{martens2010deep} for more details.

\textbf{Stochastic estimation.} With large datasets, standard CG can be slow; each iteration still goes through all $n$ training points. We use a method developed by \citet{agarwal2016second} to get an estimator that only samples a single point per iteration, which results in significant speedups.

Dropping the $\hat\theta$ subscript for clarity, let $H_j^{-1} \eqdef \sum_{i=0}^j (I - H)^i$, the first $j$ terms in the Taylor expansion of $H^{-1}$. Rewrite this recursively as $H_j^{-1} = I + (I - H)H^{-1}_{j-1}$. From the validity of the Taylor expansion, $H_j^{-1} \rightarrow H^{-1}$ as $j \rightarrow \infty$.\footnote{We assume w.l.o.g. that $\forall i, \nabla^2_\theta L(z_i, \hat\theta) \preccurlyeq I$; if this is not true, we can scale the loss down without affecting the parameters. In some cases, we can get an upper bound on $\nabla^2_\theta L(z_i, \hat\theta)$ (e.g., for linear models and bounded input), which makes this easy. Otherwise, we treat the scaling as a separate hyperparameter and tune it such that the Taylor expansion converges.}  The key is that at each iteration, we can substitute the full $H$ with a draw from any unbiased (and faster-to-compute) estimator of $H$ to form $\tilde H_j$. Since $\mathbb{E}[\tilde H_j^{-1}] = H_j^{-1}$, we still have $\mathbb{E}[\tilde H_j^{-1}] \rightarrow H^{-1}$.

In particular, we can use $\nabla^2_\theta L(z_i, \hat\theta)$, for any $z_i$, as an unbiased estimator of $H$. This gives us the following procedure:
uniformly sample $t$ points $z_{s_1}, \ldots, z_{s_t}$ from the training data;
define $\tilde H_0^{-1}v = v$; and
recursively compute $\tilde H_j^{-1}v = v + \big(I  - \nabla^2_\theta L(z_{s_j}, \hat\theta)\big) \tilde H_{j-1}^{-1}v$,
taking $\tilde H_t^{-1}v$ as our final unbiased estimate of $H^{-1}v$. We pick $t$ to be large enough such that $\tilde H_t$ stabilizes, and to reduce variance we repeat this procedure $r$ times and average results. Empirically, we found this significantly faster than CG.

We note that the original method of \citet{agarwal2016second} dealt only with generalized linear models, for which $[\nabla^2_\theta L(z_i, \hat\theta)]v$ can
be efficiently computed in $O(p)$ time. In our case, we rely on \citet{pearlmutter1994fast}'s more general algorithm for fast HVPs, described above, to achieve the same time complexity.\footnote{To increase stability, especially with non-convex models (see Section \ref{sec:nonconvex}), we can also sample a minibatch of training points at each iteration, instead of relying on a single training point.}

With these techniques, we can compute $\inflloss(z_i, z_\text{test})$ on all training points $z_i$ in $O(np + rtp)$ time; we show in Section \ref{sec:retraining} that empirically, choosing $rt = O(n)$ gives accurate results. Similarly, we can compute $\inflinput(z_i, z_\text{test}) = -\frac{1}{n} \nabla_\theta L(z_\text{test}, \hat\theta)^\top H_{\hat\theta}^{-1} \nabla_x \nabla_\theta L(z_i, \hat\theta)$ with two matrix-vector products: we first compute $s_\text{test}$, then find $s_\text{test}^\top \nabla_x \nabla_\theta L(z_i, \hat\theta)$ with the same HVP trick. These computations are easy to implement in auto-grad systems like TensorFlow \citep{abadi2015tensorflow} and Theano \citep{theano2016theano}, as users need only specify the loss; the rest is automatically handled.

\section{Validation and extensions}
\label{sec:approx}

Recall that influence functions are asymptotic approximations of leave-one-out retraining
under the assumptions that
(i) the model parameters $\hat\theta$ minimize the empirical risk,
and that (ii) the empirical risk is twice-differentiable and strictly convex. Here, we empirically show that influence functions are accurate approximations (Section \ref{sec:retraining}) that provide useful information even when these assumptions are violated (Sections \ref{sec:nonconvex}, \ref{sec:svm}).

\subsection{Influence functions vs. leave-one-out retraining}
\label{sec:retraining}

\begin{figure}[!t]
\centerline{
\includegraphics[width=\columnwidth]{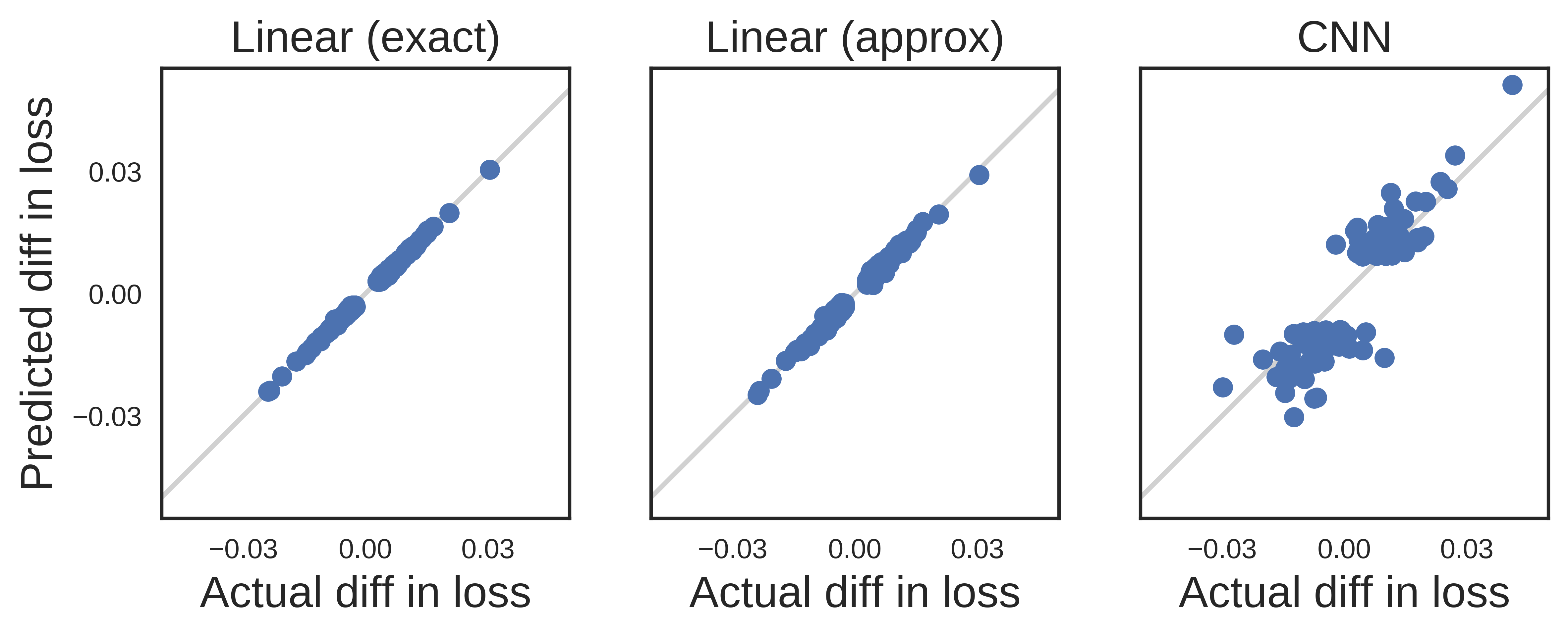}}
\vspace{-3mm}
\caption{{\bf Influence matches leave-one-out retraining.} We arbitrarily picked a wrongly-classified test point $z_\text{test}$, but this trend held more broadly. These results are from 10-class MNIST. {\bf Left:} For each of the 500 training points $z$ with largest $\big| \inflloss(z, z_\text{test}) \big|$, we plotted $-\frac{1}{n} \cdot \inflloss(z, z_\text{test})$ against the actual change in test loss after removing that point and retraining. The inverse HVP was solved exactly with CG. {\bf Mid:} Same, but with the stochastic approximation. {\bf Right:} The same plot for a CNN, computed on the 100 most influential points with CG.
For the actual difference in loss, we removed each point and retrained from $\tilde \theta$ for 30k steps.
\vspace{-7mm}
}
\label{fig:retraining}
\end{figure}

\begin{figure*}[!t]
\centerline{
\includegraphics[width=\textwidth]{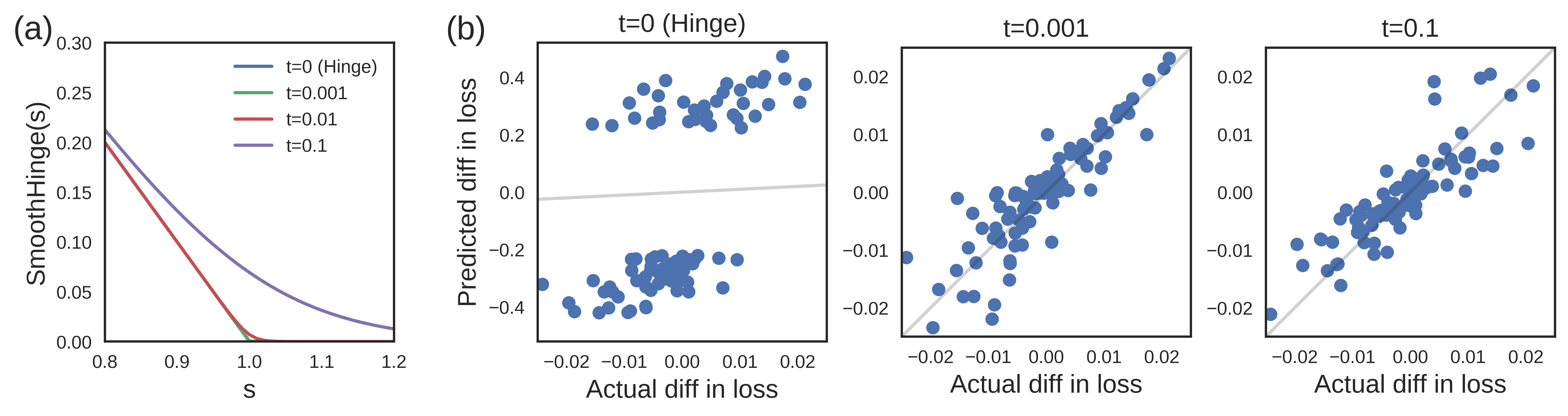}}
\vspace{-5mm}
  \caption{{\bf Smooth approximations to the hinge loss.} {\bf (a)} By varying $t$, we can approximate the hinge loss with arbitrary accuracy: the green and blue lines are overlaid on top of each other. {\bf (b)} Using a random, wrongly-classified test point, we compared the predicted vs. actual differences in loss after leave-one-out retraining on the 100 most influential training points. A similar trend held for other test points. The SVM objective is to minimize $0.005 \twonorm{w} + \frac{1}{n} \sum_i \mathrm{Hinge}(y_i w^\top x_i)$.
  {\bf Left:} Influence functions were unable to accurately predict the change, overestimating its magnitude considerably. {\bf Mid:} Using $\mathrm{SmoothHinge}(\cdot, 0.001)$ let us accurately predict the change in the hinge loss after retraining. {\bf Right:} Correlation remained high over a wide range of $t$, though it degrades when $t$ is too large. When $t = 0.001$, Pearson's R = 0.95; when $t = 0.1$, Pearson's R = 0.91.
 }
\vspace{-2mm}
\label{fig:svm}
\end{figure*}

Influence functions assume that the weight on a training point is changed by an infinitesimally small $\epsilon$.
To investigate the accuracy of using influence functions to approximate the effect of removing a training point and retraining, we compared $-\frac{1}{n} \inflloss(z, z_\text{test})$ with $L(z_\text{test}, \hat\theta_{-z}) - L(z_\text{test}, \hat\theta)$ (i.e., actually doing leave-one-out retraining). With a logistic regression model on 10-class MNIST,\footnote{We trained with L-BFGS \citep{liu1989limited}, with L$_2$ regularization of $0.01$, $n=55,000$, and $p=7,840$ parameters.}  the predicted and actual changes matched closely (Fig \ref{fig:retraining}-Left).

The stochastic approximation from \citet{agarwal2016second} was also accurate with $r=10$ repeats and $t=5,000$ iterations (Fig \ref{fig:retraining}-Mid). Since each iteration only requires one HVP $[\nabla^2_\theta L(z_i, \hat\theta)]v$, this runs quickly: in fact, we accurately estimated $H^{-1}v$ without even looking at every data point, since $n=55,000 > rt$. Surprisingly, even $r=1$ worked; while results were noisier, it was still able to identify the most influential points.

\subsection{Non-convexity and non-convergence}
\label{sec:nonconvex}
In Section \ref{sec:approach}, we took $\hat\theta$ as the global minimum. In practice, if we obtain our parameters $\tilde\theta$ by running SGD with early stopping or on non-convex objectives,
$\tilde\theta \neq \hat\theta$.
As a result, $H_{\tilde\theta}$ could have negative eigenvalues. We show that influence functions on $\tilde\theta$ still give meaningful results in practice.

Our approach is to form a convex quadratic approximation of the loss around $\tilde\theta$, i.e.,
$\tilde L(z, \theta) = L(z, \tilde\theta) + \nabla L(z, \tilde\theta)^\top (\theta - \tilde\theta) + \inv{2} (\theta - \tilde\theta)^\top (H_{\tilde\theta} + \lambda I) (\theta - \tilde\theta)$. Here, $\lambda$ is a damping term that we add if $H_{\tilde\theta}$ has negative eigenvalues; this corresponds to adding $L_2$ regularization on the parameters. We then calculate $\inflloss$
using $\tilde L$.
If $\tilde\theta$ is close to a local minimum, this is correlated with the result of taking a Newton step from $\tilde\theta$ after removing $\epsilon$ weight from $z$ (see appendix B).

We checked the behavior of $\inflloss$ in a non-convergent, non-convex setting by training a convolutional neural network for 500k iterations.\footnote{The network had 7 sets of convolutional layers with $\mathrm{tanh}(\cdot)$ non-linearities, modeled after the all-convolutional network from \citep{springenberg2014striving}. For speed, we used 10\% of the MNIST training set and only 2,616 parameters, since repeatedly retraining the network was expensive. Training was done with mini-batches of 500 examples and the Adam optimizer \citep{kingma2015adam}. The model had not converged after 500k iterations; training it for another 500k iterations, using a full training pass for each iteration, reduced train loss from 0.14 to 0.12.} The model had not converged and $H_{\tilde\theta}$ was not PD, so we added a damping term with $\lambda = 0.01$. Even in this difficult setting, the predicted and actual changes in loss were highly correlated (Pearson's R = 0.86, Fig \ref{fig:retraining}-Right).

\subsection{Non-differentiable losses}
\label{sec:svm}

What happens when the derivatives of the loss, $\nabla_\theta L$ and $\nabla_\theta^2 L$, do not exist?
In this section, we show that influence functions computed on smooth approximations to non-differentiable losses can predict the behavior of the original, non-differentiable loss under leave-one-out retraining.
The robustness of this approximation suggests that we can train non-differentiable models and swap out non-differentiable components for smoothed versions for the purposes of calculating influence.

To see this, we trained a linear SVM on the same 1s vs. 7s MNIST task in Section \ref{sec:knn}. This involves minimizing $\mathrm{Hinge}(s) = \max(0, 1-s)$; this simple piecewise linear function is similar to ReLUs, which cause non-differentiability in neural networks. We set the derivatives at the hinge to 0 and calculated $\inflloss$. As one might expect, this was inaccurate (Fig \ref{fig:svm}b-Left):
the second derivative carries no information about how close a support vector $z$ is to the hinge,
so the quadratic approximation of $L(z, \hat\theta)$ is linear,
which leads to $\inflloss(z, z_\text{test})$ overestimating the influence of $z$.

For the purposes of calculating influence, we approximated $\mathrm{Hinge}(s)$ with $\mathrm{SmoothHinge}(s,t) = t \log(1 + \exp(\frac{1-s}{t}))$, which approaches the hinge loss as $t \rightarrow 0$ (Fig \ref{fig:svm}a).
Using the same SVM weights as before, we found that calculating $\inflloss$ using $\mathrm{SmoothHinge}(s,0.001)$ closely matched the actual change due to retraining in the original $\mathrm{Hinge(s)}$ (Pearson's R = 0.95; Fig \ref{fig:svm}b-Mid) and remained accurate over a wide range of $t$ (Fig \ref{fig:svm}b-Right).

\section{Use cases of influence functions}
\label{sec:applications}

\subsection{Understanding model behavior}
\label{sec:model-behavior}

By telling us the training points ``responsible'' for a given prediction, influence functions reveal insights about how models rely on and extrapolate from the training data. In this section, we show that two models can make the same correct predictions but get there in very different ways.

\begin{figure}[h]
\centerline{
\includegraphics[width=\columnwidth]{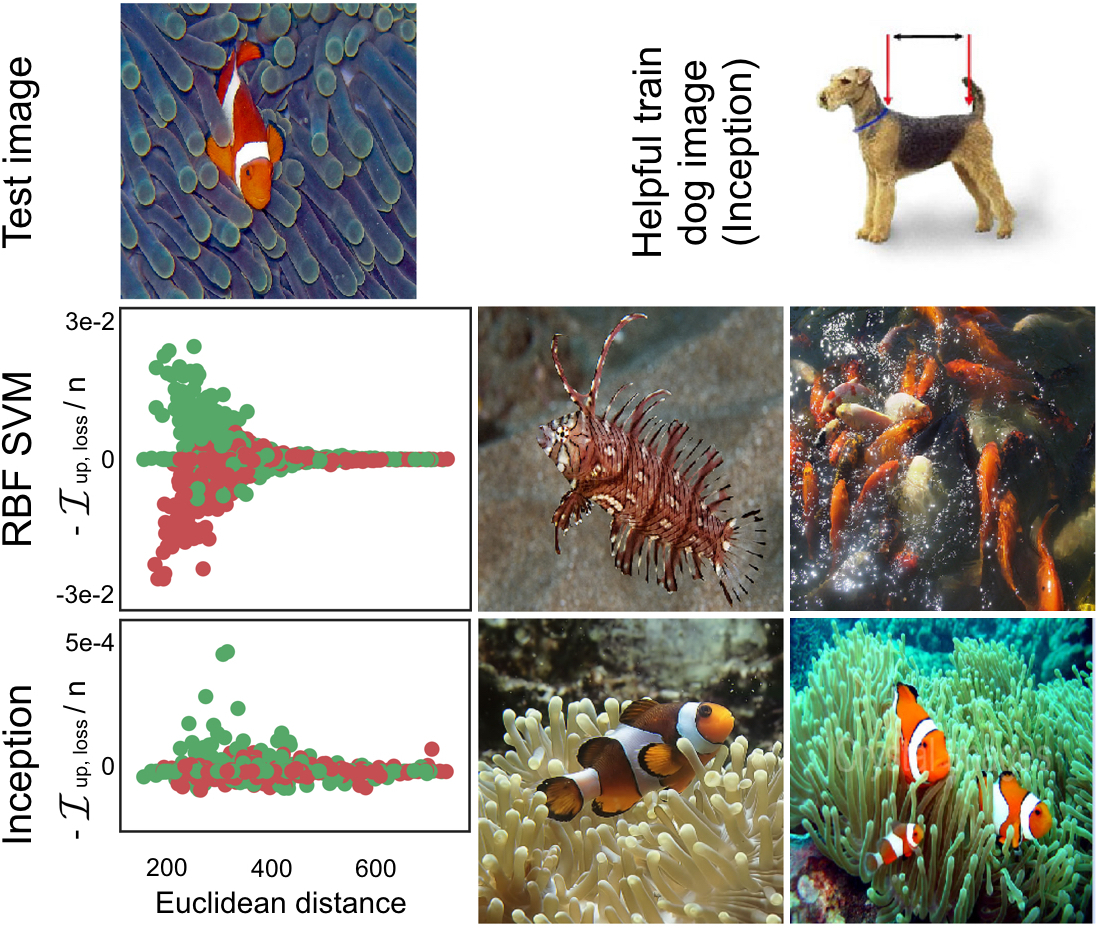}}
\vspace{-2mm}
\caption{{\bf Inception vs. RBF SVM.} {\bf Bottom left:} $-\inflloss(z, z_\text{test})$ vs. $\twonorm{z - z_\text{test}}$. Green dots are fish and red dots are dogs. {\bf Bottom right:} The two most helpful training images, for each model, on the test. {\bf Top right:} An image of a dog in the training set that helped the Inception model correctly classify the test image as a fish.}
\vspace{-4mm}
\label{fig:rbf-comparison}
\end{figure}

We compared (a) the state-of-the-art Inception v3 network \citep{szegedy2016rethinking} with all but the top layer frozen\footnote{We used pre-trained weights from Keras \citep{chollet2015keras}.} and (b) an SVM with an RBF kernel on a dog vs. fish image classification dataset we extracted from ImageNet \citep{russakovsky2015imagenet}, with 900 training examples for each class. Freezing neural networks in this way is not uncommon in computer vision and is equivalent to training a logistic regression model on the bottleneck features \citep{donahue2014decaf}. We picked a test image  both models got correct (Fig \ref{fig:rbf-comparison}-Top) and used $\mathrm{SmoothHinge}(\cdot, 0.001)$ to compute the influence for the SVM.

As expected, $\inflloss$ in the RBF SVM varied inversely with raw pixel distance, with training images far from the test image in pixel space having almost no influence; the Inception influences were much less correlated with distance in pixel space (Fig \ref{fig:rbf-comparison}-Left). Looking at the two most helpful images (most positive $-\inflloss$) for each model in Fig \ref{fig:rbf-comparison}-Right, we see that the Inception network picked on the distinctive characteristics of clownfish, whereas the RBF SVM pattern-matched training images superficially.

Moreover, in the RBF SVM, fish (green points) close to the test image were mostly helpful, while dogs (red) were mostly harmful, with the RBF acting as a soft nearest neighbor function (Fig \ref{fig:rbf-comparison}-Left). In contrast, in the Inception network, fish and dogs could be helpful or harmful for correctly classifying the test image as a fish; in fact, the 5th most helpful training image was a dog that, to the model, looked very different from the test fish (Fig \ref{fig:rbf-comparison}-Top).

\subsection{Adversarial training examples}
\label{sec:dataset-poisoning}

\begin{figure*}
  \vspace{-1mm}
  \begin{minipage}[c]{0.78\textwidth}
    \includegraphics[width=\textwidth]{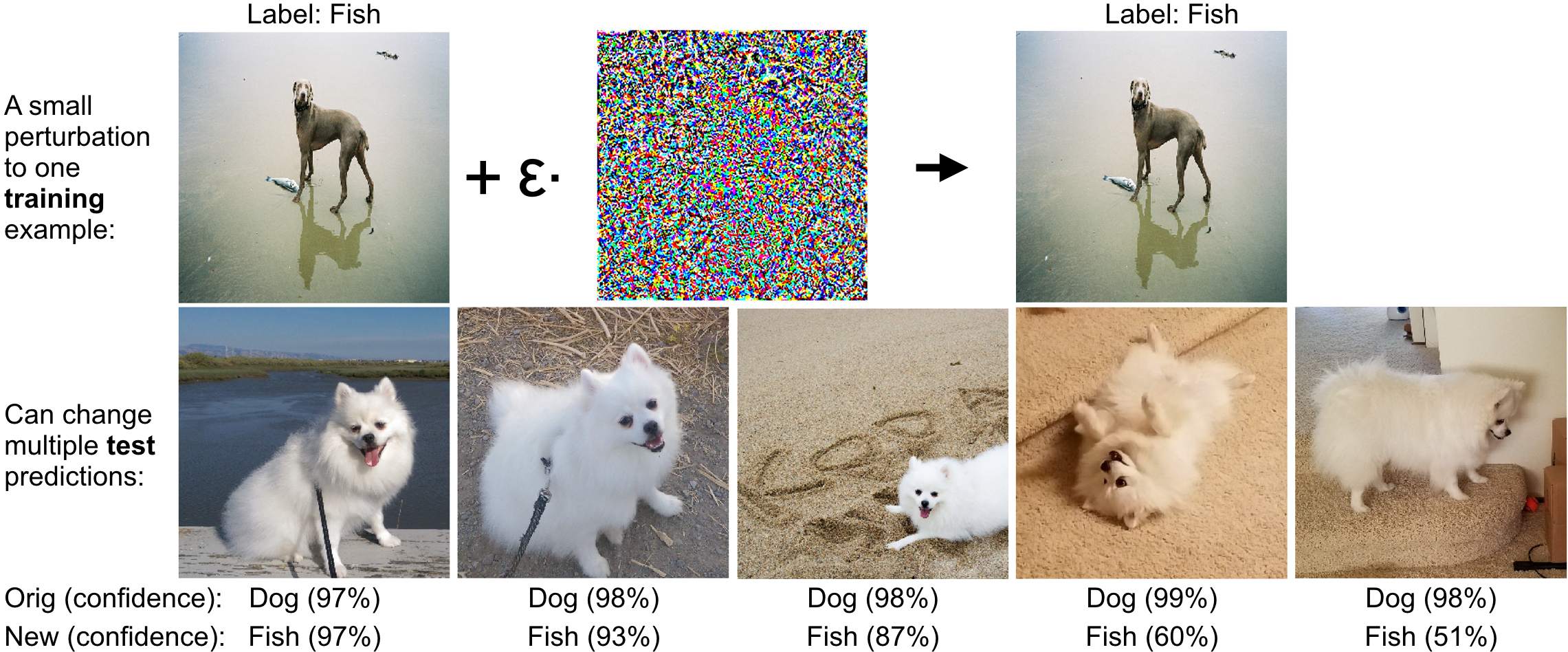}
  \end{minipage}\hfill
  \begin{minipage}[c]{0.20\textwidth}
    \caption{{\bf Training-set attacks.}
    We targeted a set of 30 test images featuring the first author's dog in a variety of poses and backgrounds. By maximizing the average loss over these 30 images,
    we found a visually-imperceptible change to the particular training image (shown on top) that flipped predictions on 16 test images.
    }
    \label{fig:attack}
  \end{minipage}
  \vspace{-3mm}
\end{figure*}

In this section, we show that models that place a lot of influence on a small number of points can be vulnerable to training input perturbations, posing a serious security risk in real-world ML systems where attackers can influence the training data \citep{huang2011adversarial}. Recent work has generated adversarial \emph{test} images that are visually indistinguishable from real test images but completely fool a classifier \citep{goodfellow2015explaining, moosavi2016deepfool}. We demonstrate that influence functions can be used to craft adversarial \emph{training} images that are similarly visually-indistinguishable and can flip a model's prediction on a separate test image. To the best of our knowledge, this is the first proof-of-concept that visually-indistinguishable training attacks can be executed on otherwise highly-accurate neural networks.

The key idea is that $\inflinput(z, z_\text{test})$ tells us how to modify training point $z$ to most increase the loss on $z_\text{test}$. Concretely,
for a target test image $z_\text{test}$, we can construct $\tilde z_i$, an adversarial version of a training image $z_i$, by initializing $\tilde z_i := z_i$ and then iterating $\tilde z_i := \Pi(\tilde z_i + \alpha \sign(\inflinput(\tilde z_i, z_\text{test})))$, where $\alpha$ is a step size and $\Pi$ projects onto the set of valid images that share the same 8-bit representation with $z_i$. After each iteration, we retrain the model. This is an iterated, training-set analogue of the methods used by, e.g., \citet{goodfellow2015explaining, moosavi2016deepfool} for test-set attacks.

We tested these adversarial training perturbations on the same Inception network on dogs vs. fish from Section \ref{sec:model-behavior}, choosing this pair of animals to provide a stark contrast between the classes. We set $\alpha=0.02$ and ran the attack for 100 iterations on each test image.
As before, we froze all but the top layer for training; note that computing $\inflinput$ still involves differentiating through the entire network.
Originally, the model correctly classified 591 / 600 test images.
For each of these 591 test images, considered separately, we tried to find a visually-indistinguishable perturbation (i.e., same 8-bit representation) to a single training image, out of 1,800 total training images, that would flip the model's prediction.
We were able to do this on 335 (57\%) of the 591 test images. If we perturbed 2 training images for each test image, we could flip predictions on 77\% of the 591 test images; and if we perturbed 10 training images, we could flip all but 1 of the 591.
The above results are from attacking each test image separately, i.e., we use a different training set to attack each test image.
We next tried to attack multiple test images simultaneously by increasing their average test loss,
and found that single training image perturbations could simultaneously flip multiple test predictions as well (Fig \ref{fig:attack}).

We make three observations about these attacks.
First, though the change in pixel values is small, the change in the final Inception feature layer is significantly larger:
in pixel space and using $L_2$ distance, the training values change by less than $1\%$ of the mean distance of a training point to the class centroid,
whereas in Inception feature space, the change is on the same order as the mean distance.
Second, the attack tries to perturb the training example in a direction of low variance, causing the model to overfit in that direction
and consequently incorrectly classify the test images; we expect the attack to become harder as the number of training examples grows.
Third, ambiguous or mislabeled training images are effective points to attack, since the model has low confidence and thus high loss on them,
making them highly influential (recall Section \ref{sec:knn}).
For example, the image in Fig \ref{fig:attack} contains both a dog and a fish and is highly ambiguous; as a result, it is the training
example that the model is least confident on (with a confidence of 77\%, compared to the next lowest confidence of 90\%).

This attack is mathematically equivalent to the gradient-based training set attacks explored by \citet{biggio2012poisoning, mei2015teaching} and others in the context of different models.\citet{biggio2012poisoning} constructed a dataset poisoning attack against a linear SVM on a two-class MNIST task, but had to modify the training points in an obviously distinguishable way to be effective. Measuring the magnitude of $\inflinput$ gives model developers a way of quantifying how vulnerable their models are to training-set attacks.

\subsection{Debugging domain mismatch}
\label{sec:covariate-shift}
Domain mismatch --- where the training distribution does not match the test distribution --- can cause models with high training accuracy to do poorly on test data \citep{ben2010theory}. We show that influence functions can identify the training examples most responsible for the errors, helping model developers identify domain mismatch.

As a case study, we predicted whether a patient would be readmitted to a hospital. Domain mismatches are common in biomedical data; for example, different hospitals can serve very different populations, and readmission models trained on one population can do poorly on another \citep{kansagara2011risk}. We used logistic regression to predict readmission with a balanced training dataset of 20K diabetic patients from 100+ US hospitals, each represented by 127 features \citep{strack2014impact}.\footnote{Hospital readmission was defined as whether a patient would be readmitted within the next 30 days. Features were demographic (e.g., age, race, gender), administrative (e.g., length of hospital stay), or medical (e.g., test results). }

3 out of the 24 children under age 10 in this dataset were re-admitted. To induce a domain mismatch, we filtered out 20 children who were not re-admitted, leaving 3 out of 4 re-admitted. This caused the model to wrongly classify many children in the test set. Our aim is to identify the 4 children in the training set as being ``responsible'' for these errors.

As a baseline, we tried the common practice of looking at the learned parameters $\hat\theta$ to see if the indicator variable for being a child was obviously different. However, this did not work: 14/127 features had a larger coefficient.

Picking a random child $z_\text{test}$ that the model got wrong, we calculated $-\inflloss(z_i, z_\text{test})$ for each training point $z_i$. This clearly highlighted the 4 training children, each of whom were 30-40 times as influential as the next most influential examples. The 1 child in the training set who was not readmitted had a very positive influence, while the other 3 had very negative influences. Calculating $\inflinput$ on these 4 children showed that a change in the `child' indicator variable had by far the largest effect on $\inflloss$.

\subsection{Fixing mislabeled examples}

Labels in the real world are often noisy, especially if crowdsourced \citep{frenay2014classification}, and can even be adversarially corrupted, as in Section \ref{sec:dataset-poisoning}. Even if a human expert could recognize wrongly labeled examples, it is impossible in many applications to manually review all of the training data. We show that influence functions can help human experts prioritize their attention, allowing them to inspect only the examples that actually matter.

The key idea is to flag the training points that exert the most influence on the model. Because we do not have access to the test set, we measure the influence of $z_i$ with $\inflloss(z_i, z_i)$, which approximates the error incurred on $z_i$ if we remove $z_i$ from the training set.

\begin{figure}[!ht]
\centerline{
\includegraphics[width=1.05\columnwidth]{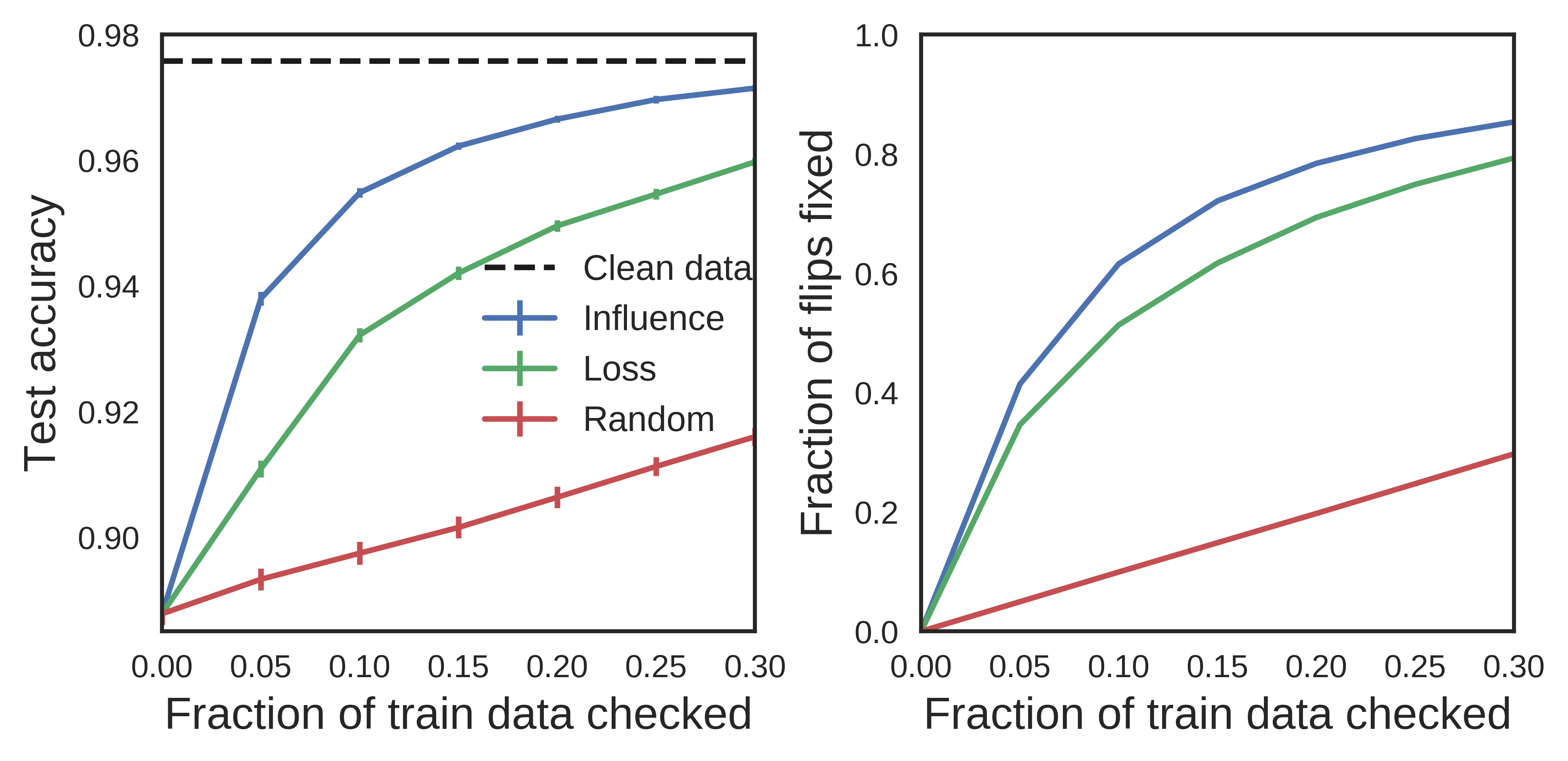}}
\vspace{-2mm}
  \caption{{\bf Fixing mislabeled examples.} Plots of how test accuracy (left) and the fraction of flipped data detected (right) change with the fraction of train data checked, using different algorithms for picking points to check. Error bars show the std. dev. across 40 repeats of this experiment, with a different subset of labels flipped in each; error bars on the right are too small to be seen. These results are on the Enron1 spam dataset \citep{metsis2006spam}, with 4,147 training and 1,035 test examples; we trained logistic regression on a bag-of-words representation of the emails.}
\vspace{-4mm}
\label{fig:spam}
\end{figure}

Our case study is email spam classification, which relies on user-provided labels and is also vulnerable to adversarial attack \citep{biggio2011label}. We flipped the labels of a random 10\% of the training data and then simulated manually inspecting a fraction of the training points, correcting them if they had been flipped. Using influence functions to prioritize the training points to inspect allowed us to repair the dataset (Fig \ref{fig:spam}, blue) without checking too many points, outperforming the baselines of
checking points with the highest train loss (Fig \ref{fig:spam}, green) or at random (Fig \ref{fig:spam}, red).
No method had access to the test data.
\label{sec:fix}

\section{Related work}
\label{sec:related}

The use of influence-based diagnostics originated in statistics in the 70s, driven by the seminal papers of \citet{hampel1974influence} and \citet{jaeckel1972infinitesimal} (where it was called the infinitesimal jackknife). It was further developed in the book by \citet{hampel1986robust} and many other contemporary papers \citep{cook1977detection, cook1980influence, pregibon1981logistic, cook1982residuals}.
Earlier work focused on removing training points from linear models, with later work extending this to more general models and a wider variety of perturbations \citep{hampel1986robust, cook1986assessment, thomas1990assessing, chatterjee1986influential, wei1998generalized}.
Prior work mostly focused on experiments with small datasets, e.g., $n=24$ and $p=10$ in \citet{cook1980influence}, and thus paid special attention to exact solutions, or if not possible, characterizations of the error terms.

Influence functions have not been used much in the ML literature, with some exceptions. \citet{christmann2004robustness, debruyne2008model, liu2014efficient} use influence functions to study model robustness and to do fast cross-validation in kernel methods. \citet{wojnowicz2016sketching} use matrix sketching to estimate Cook's distance, which is closely related to influence; they focus on prioritizing training points for human attention and derive methods specific to generalized linear models.
\citet{kabra2015understanding} define a different notion of influence that is specialized to finite hypothesis classes.

As noted in Section \ref{sec:dataset-poisoning}, our training-set attack is mathematically equivalent to an approach first explored by \citet{biggio2012poisoning} in the context of SVMs, with follow-up work extending the framework and applying it to linear and logistic regression \citep{mei2015teaching}, topic modeling \citep{mei2015security}, and collaborative filtering \citep{li2016data}. These papers derived the attack directly from the KKT conditions without considering influence, though for continuous data, the end result is equivalent. Influence functions additionally let us consider attacks on discrete data (Section \ref{sec:approach-input}), but we have not tested this empirically.
Our work connects the literature on training-set attacks with work on ``adversarial examples'' \cite{goodfellow2015explaining, moosavi2016deepfool}, visually-imperceptible perturbations on test inputs.

In contrast to training-set attacks, \citet{cadamuro2016debugging} consider the task of taking an incorrect test prediction and finding a small subset of training data such that changing the labels on this subset makes the prediction correct. They provide a solution for OLS and Gaussian process models when the labels are continuous. Our work with influence functions allow us to solve this problem in a much larger range of models and in datasets with discrete labels.

\section{Discussion}

We have discussed a variety of applications, from creating training-set attacks to debugging models and fixing datasets. Underlying each of these applications is a common tool, influence functions, which are based on a simple idea---we can better understand model behavior by looking at how it was derived from its training data.

At their core, influence functions measure the effect of local changes: what happens when we upweight a point by an infinitesimally-small $\epsilon$? This locality allows us to derive efficient closed-form estimates, and as we show, they can be surprisingly effective.  However, we might want to ask about more global changes, e.g., how does a subpopulation of patients from this hospital affect the model? Since influence functions depend on the model not changing too much, how to tackle this is an open question.

It seems inevitable that high-performing, complex, black-box models will become increasingly prevalent and important. We hope that the approach presented here---of looking at the model through the lens of the training data---will become a standard part of the toolkit of developing, understanding, and diagnosing machine learning.

\section*{Reproducibility}
The code and data for replicating our experiments is available on GitHub \url{http://bit.ly/gt-influence} and Codalab \url{http://bit.ly/cl-influence}.

\section*{Acknowledgements}
We thank Jacob Steinhardt, Zhenghao Chen, and Hongseok Namkoong for helpful discussions and comments. We are also grateful to Doug Martin, Swee Keat Lim, and Teresa Yeo for finding typos and omissions in a previous version of the manuscript. This work was supported by a Future of Life Research Award and a Microsoft Research Faculty Fellowship.

\setcounter{section}{0}
\renewcommand{\thesection}{\Alph{section}}
\section{Deriving the influence function $\inflparams$}
For completeness, we provide a standard derivation of the influence function $\inflparams$ in the context of loss minimization (M-estimation).
This derivation is based on asymptotic arguments and is not fully rigorous;
see \citet{vaart98asymptotic} and other statistics textbooks for a more thorough treatment.

Recall that $\hat\theta$ minimizes the empirical risk:
\begin{align}
  R(\theta) \eqdef \frac{1}{n} \sum_{i=1}^n L(z_i,\theta).
\end{align}
We further assume that $R$ is twice-differentiable and strongly convex in $\theta$,
i.e.,
\begin{align}
  \label{eqn:defineHessian}
  H_{\hat\theta} \eqdef \nabla^2 R(\hat\theta) = \frac{1}{n} \sum_{i=1}^n \nabla^2_\theta L(z_i, \hat\theta)
\end{align}
exists and is positive definite. This guarantees the existence of $H_{\hat\theta}^{-1}$, which we will use in the subsequent derivation.

The perturbed parameters $\hat\theta_{\epsilon,z}$ can be written as
\begin{align}
  \label{eqn:perturbedOpt}
  \hat\theta_{\epsilon,z} = \arg\min_{\theta \in \Theta} \pc{R(\theta) + \epsilon L(z, \theta)}.
\end{align}

Define the parameter change $\Delta_\epsilon = \hat\theta_{\epsilon,z} - \hat\theta$,
and note that, as $\hat\theta$ doesn't depend on $\epsilon$, the quantity we seek to compute can be written in terms of it:
\begin{align}
  \label{eqn:thetaDelta}
  \frac{d\hat\theta_{\epsilon,z}}{d\epsilon} = \frac{d\Delta_\epsilon}{d\epsilon}.
\end{align}

Since $\hat\theta_{\epsilon,z}$ is a minimizer of \refeqn{perturbedOpt}, let us examine its first-order optimality conditions:
\begin{align}
  0 = \nabla R(\hat\theta_{\epsilon,z}) + \epsilon \nabla L(z, \hat\theta_{\epsilon,z}).
\end{align}

Next, since $\hat\theta_{\epsilon,z} \cv \hat\theta$ as $\epsilon \cv 0$,
we perform a Taylor expansion of the right-hand side:
\begin{align}
  0 \approx & \pb{\nabla R(\hat\theta) + \epsilon \nabla L(z, \hat\theta)} + \\
  & \pb{\nabla^2 R(\hat\theta) + \epsilon \nabla^2 L(z, \hat\theta)} \Delta_\epsilon, \nonumber
\end{align}
where we have dropped $o(\|\Delta_\epsilon\|)$ terms.

Solving for $\Delta_\epsilon$, we get:
\begin{align}
  \Delta_\epsilon \approx & -\pb{\nabla^2 R(\hat\theta) + \epsilon \nabla^2 L(z, \hat\theta)}^{-1} \\
  & \pb{\nabla R(\hat\theta) + \epsilon \nabla L(z, \hat\theta)}. \nonumber
\end{align}
Since $\hat\theta$ minimizes $R$, we have $\nabla R(\hat\theta) = 0$.
Dropping $o(\epsilon)$ terms, we have
\begin{align}
  \Delta_\epsilon \approx & -\nabla^2 R(\hat\theta)^{-1} \nabla L(z, \hat\theta) \epsilon.
\end{align}

Combining with \refeqn{defineHessian} and \refeqn{thetaDelta},
we conclude that:
\begin{align}
  \frac{d\hat\theta_{\epsilon,z}}{d\epsilon}\Bigr|_{\substack{\epsilon = 0}} &= -H_{\hat\theta}^{-1} \nabla L(z, \hat\theta) \\
  &\eqdef \inflparams(z).
\end{align}

\section{Influence at non-convergence}
\label{section:non-convex}
Consider a training point $z$. When the model parameters $\tilde\theta$ are close to but not at a local minimum, $\inflparams(z)$ is approximately equal to a constant (which does not depend on $z$) plus the change in parameters after upweighting $z$ and then taking a single Newton step from $\tilde\theta$. The high-level idea is that even though the gradient of the empirical risk at $\tilde\theta$ is not 0, the Newton step from $\tilde\theta$ can be decomposed into a component following the existing gradient (which does not depend on the choice of $z$) and a second component responding to the upweighted $z$ (which $\inflparams(z)$ tracks).

Let $g \eqdef \frac{1}{n} \sum_{i=1}^n \nabla_\theta L(z_i, \tilde\theta)$ be the gradient of the empirical risk at $\tilde\theta$; since $\tilde\theta$ is not a local minimum, $g \neq 0$. After upweighting $z$ by $\epsilon$, the gradient at $\tilde\theta$ goes from $g \mapsto g  + \epsilon \nabla_\theta L(z, \tilde\theta)$, and the empirical Hessian goes from $H_{\tilde\theta} \mapsto H_{\tilde\theta} + \epsilon  \nabla_\theta^2 L(z, \tilde\theta)$. A Newton step from $\tilde\theta$ therefore changes the parameters by:
\begin{align}
N_{\epsilon, z} \eqdef -&\left[ H_{\tilde\theta} + \epsilon  \nabla_\theta^2 L(z, \tilde\theta)  \right]^{-1}
\left[ g  +
\epsilon  \nabla_\theta L(z, \tilde\theta) \right].
\end{align}

Ignoring terms in $\epsilon g$, $\epsilon^2$, and higher,
we get $N_{\epsilon, z} \approx -H_{\tilde\theta}^{-1}\left(g + \epsilon \nabla_\theta L(z, \tilde\theta)\right)$.
Therefore, the actual change due to a Newton step $N_{\epsilon,z}$ is equal to
a constant $-H_{\tilde\theta}^{-1} g$ (that doesn't depend on $z$)
plus $\epsilon$ times $\inflparams(z) = -H_{\tilde\theta}^{-1} \nabla_\theta L(z, \tilde\theta)$
(which captures the contribution of $z$).

\bibliography{refdb/all}
\bibliographystyle{icml2017}

\end{document}